# Fast vehicle detection algorithm based on lightweight YOLO7-tiny

Bo Li[a], YiHua Chen[a], Hao Xu [a, *], and Fei Zhong[a]
[a] School of Mechanical Engineering, Hubei University of Technology, Wuhan 430000, China;
* Correspondence to: School of Mechanical Engineering, Hubei University of Technology, 28 NanLi Road, HongShan District, Wuhan, China.
E-mail addresses: Hao Xu (haoxu42@outlook.com), Bo Li (libo@hbut.edu.cn)

**Abstract:** The swift and precise detection of vehicles plays a significant role in intelligent transportation systems. Current vehicle detection algorithms encounter challenges of high computational complexity, low detection rate, and limited feasibility on mobile devices. To address these issues, this paper proposes a lightweight vehicle detection algorithm based on YOLOv7-tiny (You Only Look Once version seven) called Ghost-YOLOv7. The width of model is scaled to 0.5 and the standard convolution of the backbone network is replaced with Ghost convolution to achieve a lighter network and improve the detection speed; then a self-designed Ghost bi-directional feature pyramid network (Ghost-BiFPN) is embedded into the neck network to enhance feature extraction capability of the algorithm and enriches semantic information; and a Ghost Decouoled Head (GDH) is employed for accurate prediction of vehicle location and species; finally, a coordinate attention mechanism is introduced into the output layer to suppress environmental interference. The WIoU loss function is employed to further enhance the detection accuracy. Ablation experiments results on the PASCAL VOC dataset demonstrate that Ghost-YOLOv7 outperforms the original YOLOv7-tiny model. It achieving a 29.8% reduction in computation, 37.3% reduction in the number of parameters, 35.1% reduction in model weights, 1.1% higher mean average precision (mAP), the detection speed is higher 27FPS compared with the original algorithm. Ghost-YOLOv7 was also compared on KITTI and BIT-vehicle datasets as well, and the results show that this algorithm has the overall best performance.
**Keywords:** vehicle detection; lightweight; Ghost-YOLOv7; deep learning

## 1、Introduction

Nowadays, the pursuit of safety and comfort drives the development of autonomous driving technology. Self-driving car is a revolutionary tool for road traffic and a significant indicator of human progress in this new era. Vehicle detection is one of the most critical technologies in autonomous driving research. [1] Vehicle detection methods can be broadly classified into traditional and deep learning-based approaches.

Earlier researchers identified features such as edge information, color, and symmetry of images to detect vehicles. Some classical algorithms such as Scale Invariant Feature Transform (SIFT), Histogram of Oriented Gradients (HOG), and others approaches [2-4].

Tsai et al. [5] proposed a vehicle detection approach by identifying color and edges to detect vehicles from static images. Similarly, Chen et al. [6] developed a vision-based system for daytime brake light detection that extracted the symmetry verification of the headlight at night and combined brightness and radial symmetry to detect the brake light, which is helpful in identifying vehicles. As technology progressed, Support Vector Machine (SVM) and AdaBoost classifiers are widely used due to their superior performance. Cheng et al. [7] proposed a cascaded classifier by combining AdaBoost and SVM, the classifier could extract and classifier candidate regions using a fixed-size window. Satzoda et al. [8] constructed a multipart-based vehicle detection algorithm that utilized active learning and symmetry, it employed Haar features to detect fully and partially visible rear views of vehicles. Razalli et al. [9] presented a novel approach that combined Hue Saturation and Value (HSV) color segmentation and SVM to detect moving emergency vehicles in traffic cameras. Most of the above approaches use manual feature selection to design and train classifiers based on specific detection objects. However, these methods usually extract overly homogeneous features and cannot accurately detect objects in complex traffic environments.

With the continuous advancement of deep learning technology, vehicle detection methods based on deep learning have emerged as a popular area of research. This deep learning-based vehicle detection approach can be divided into Convolutional Neural Networks (CNN) based method [10] and Vision Transformer (ViT) based method. [11-12] Vision Transformer has become popular in recent years due to its higher detection accuracy on the expense of detection speed. Sun et al. [13] introduced an effective vehicle detection method based on the swin transformer algorithm for vehicle detection in fuzzy scenes. Experimental results show that the algorithm provides good detection results in different datasets and scenarios and has better robustness and accuracy than other common vehicle detection algorithms, indicating that the swin transformer effectively enhanced the performance of fuzzy backgrounds. Deshmukh et al. [14] proposed a swin transformer-based vehicle detection method for the lack of vehicle detection accuracy in a disordered traffic environment. The technique used BiFPN to enhance the feature fusion capability. It employed a fully connected vehicle detection head to improve the matching relationship between vehicle size and BiFPN, effectively improving vehicle detection accuracy and robustness on various datasets.

CNN-based approaches for vehicle detection are typically faster, cheaper, and simpler to deploy models than ViT-based ones. Arora et al. [15] used Faster R-CNN to detect vehicles in different daytime, nighttime, sunny and rainy conditions and achieved satisfactory results. Satyanarayana et al. [16] used spatiotemporal information obtained from CNN graphs for unsupervised vehicle detection, to approximate and classify the width and length of the vehicles, and the method achieved satisfactory accuracy. Zakaria et al. [17] used a lightweight CNN model to eliminate background of images. This method optimizes the number of convolutional operations performed by the CNN module to reduce the complexity of the model and maintain accuracy. The YOLO (You Only Look Once) series are the most popular algorithms among CNN-based one-stage detectors in recent years. Zhao et al. [18] proposed a deep learning-based vehicle detection method to improve detection performance by improving YOLOv4. The new method suppressed irrelevant information through an attention mechanism, and modified the neck network during downsampling to enhance feature extraction. Dong et al. [19] proposed a lightweight YOLOv5 vehicle detection model, by adding C3Ghost and Ghost modules to the neck network to reduce the weight of the model, adding CBAM modules to the backbone network to improve the detection accuracy, and employing the CIoU loss function to speed up the regression, The new model achieved good results on the PASCAL VOC and MS COCO datasets. Bie et al. [20] proposed the YOLOv5n-L model on the basis of YOLOv5n to reduce model weight. The improvements included the replacement of C3 module with depth-separable convolution and C3Ghost module to reduce parameters, combination of Squeeze-and-Excitation (SE) attention mechanism to improve detection accuracy, and employment of BiFPN for multi-scale feature fusion to enrich feature information.

However, these methods cannot achieve fast, precise detection and easy delployment simultaneously. In this paper, a lightweight YOLOv7-tiny vehicle detection model named Ghost-YOLOv7 is proposed. The model uses an efficient and straightforward structure to achieve accurate real-time detection of objects. The main contributions of this paper are as follows:

(1) The width factor of the YOLOv7-tiny network is scaled to 0.5, and Ghost convolution is used to replace the standard convolution of the backbone network to achieve a lighter model and improve the detection speed.

(2) A Ghost-BiFPN neck network is designed to enhance the feature extraction capability of the network and enrich the network information.

(3) A lightweight Ghost Decoupled Head is embedded to make the classification and localization of detection heads more focused on the information and speed up the convergence.

(4) A coordinate attention mechanism is used to improve the focus on the vital information of the vehicle detection task, and a WIoU loss function is introduced to enhance the detection accuracy.

The remainder of this paper is organized as follows: Section 2 describes the principle of YOLOv7-tiny and the details of the Ghost-YOLOv7 model. Section 3 describes the ablation experiments and the comparative analysis under different data sets. Section 4 summarizes the work of this paper.

## 2、Method

## 2.1. YOLOv7-tiny architecture

YOLOv7 [21] is one popular one-stage object detection networks. YOLOv7-tiny is the model of smallest scale in the YOLOv7 series, its backbone network consists of ELAN structures and Downsampling modules. The input image is firstly passed through the backbone network to generate three feature maps of different sizes, which passed into the neck network to further enhance the feature fusion capability in the second stage, and finally passed through the detection head to get the final detection result. Although the

YOLOv7-tiny model shows great detection accuracy and speed in lightweight models, it still has some insufficience as follows:

(1) There is still space for improvement in the number of parameters and computational effort of the YOLOv7-tiny model.

(2) The coupled head of YOLOv7-tiny does not have enough semantic information from the feature map when detecting small objects with low resolution, resulting in poor detection of small objects.

(3) YOLOv7-tiny is vulnerable to missed and false detections in complex and variable environment.

Thus, these insufficiencies prevent its widespread use in practical application in vehicle detection.

## 2.2. Ghost-YOLOv7 architecture

According to the shortcomings of the YOLOv7-tiny, this paper proposed an improved model named Ghost-YOLOv7. The structure of Ghost-YOLOv7 is shown in Figure 1 and Figure 2. Firstly, the width factor of the original YOLOv7-tiny network is tuned from 1.0 to 0.5 and the standard convolution in the backbone network is replaced with Ghost convolution to achieve a lightweight network; secondly, a self-designed Ghost-BiFPN neck network is embedded to enhance the feature extraction capability of the network; then, a light decoupling head is employed to improve the small object detection capability; finally, an attention mechanism and Wise-IoU loss function are utilized to further improve the detection accuracy. A detailed explanation of each module is shown below:

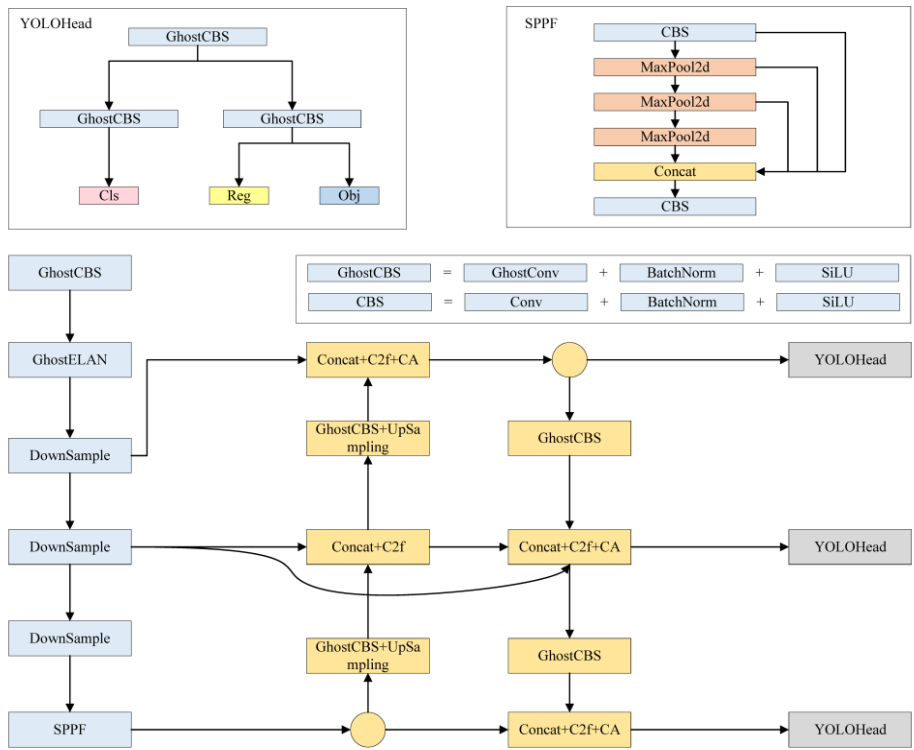

Figure 1 Framework of Ghost-YOLOv7

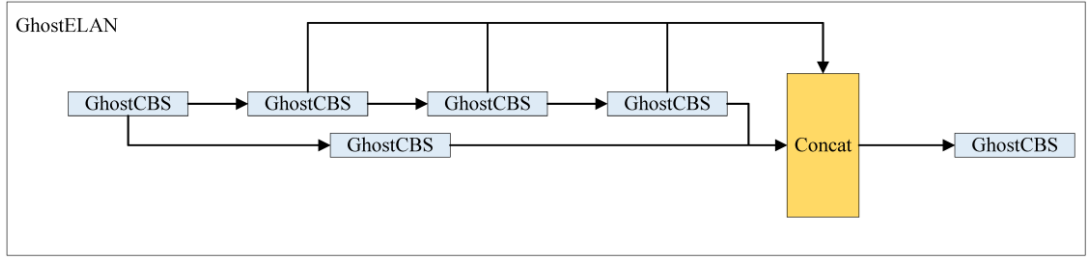

(a) GhostELAN

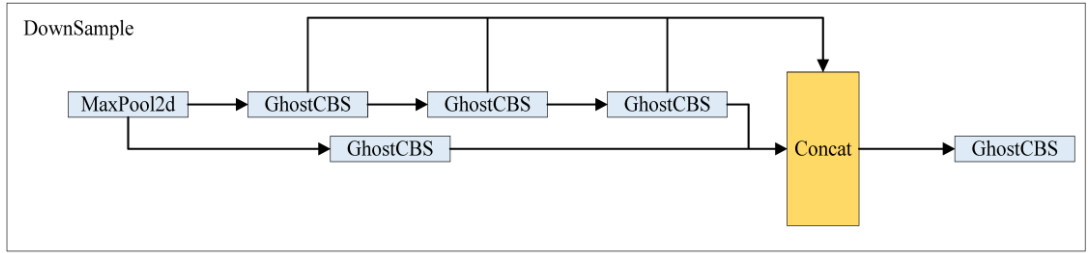

(b) DownSample

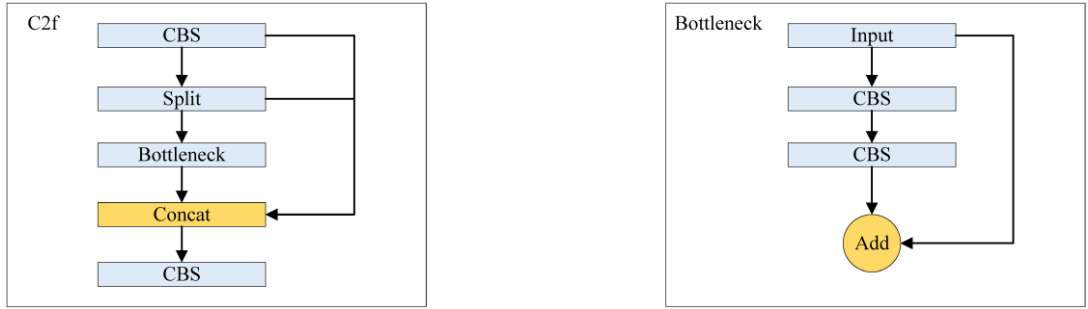

(c) C2f and Bottleneck

Figure 2 The structure of some modules in Ghost-YOLOv7

## 2.2.1. GhostConv

GhostNet [22] proposes a new Ghost module to generate more feature maps through cheaper operations. In the standard convolution operation, once the input data $X \in R^{h \times w \times c}$ is given, where $w$ and $h$ represent the width and height of the input data and $c$ is the number of input channels. The operation of the convolution layer to generate $m$ feature maps is defined as formula 1:

$$O = X * f + b \qquad (1)$$

where $b$ is the bias item, $*$ denotes the convolution operation, $O \in R^{h' \times w' \times m}$ denotes the output feature map for $m$ channels, and $f \in R^{c \times k \times k \times m}$ is the convolution filter. Furthermore, $w'$ and $h'$ are the width and height of the output data and $k \times k$ denotes the kernel size of convolution filters $f$. The standard convolution can be calculated in formula 2:

$$m \cdot w' \cdot h' \cdot c \cdot k \cdot k \tag{2}$$

Since the number of channels and filters in a network is usually large, it may lead to a massive amount of computation for standard convolution. The Ghost convolution is capable to reduce computational costs by conduct inexpensive linear operations. The size of a linear operator kernel is set to $l \times l$, suppose n feature maps are obtained by the original method then $m = n \cdot s$, $s$ represents the corresponding redundant features and $s \ll c$. The operation process of Ghost convolution is shown in Figure 3, which is specific for its identity mapping, and the actual amount of transformation is calculated in formula 3:

$$n \cdot (s-1) = \frac{m}{s} \cdot (s-1) \tag{3}$$

The computational volume of Ghost convolution can be expressed as:

$$\frac{m}{s} \cdot h' \cdot w' \cdot c \cdot k \cdot k + (s-1) \cdot \frac{m}{s} \cdot h' \cdot w' \cdot l \cdot l \tag{4}$$

The computational saving of using Ghost convolution instead of standard convolution is:

$$\begin{aligned} &\frac{\frac{m}{s} \cdot h' \cdot w' \cdot c \cdot k \cdot k + (s-1) \cdot \frac{m}{s} \cdot h' \cdot w' \cdot l \cdot l}{m \cdot h' \cdot w' \cdot c \cdot k \cdot k} \\ &= \frac{\frac{1}{s} \cdot c \cdot k \cdot k + \frac{s-1}{s} \cdot l \cdot l}{c \cdot k \cdot k} \approx \frac{s+c-1}{s \cdot c} \approx \frac{1}{s} \end{aligned} \tag{5}$$

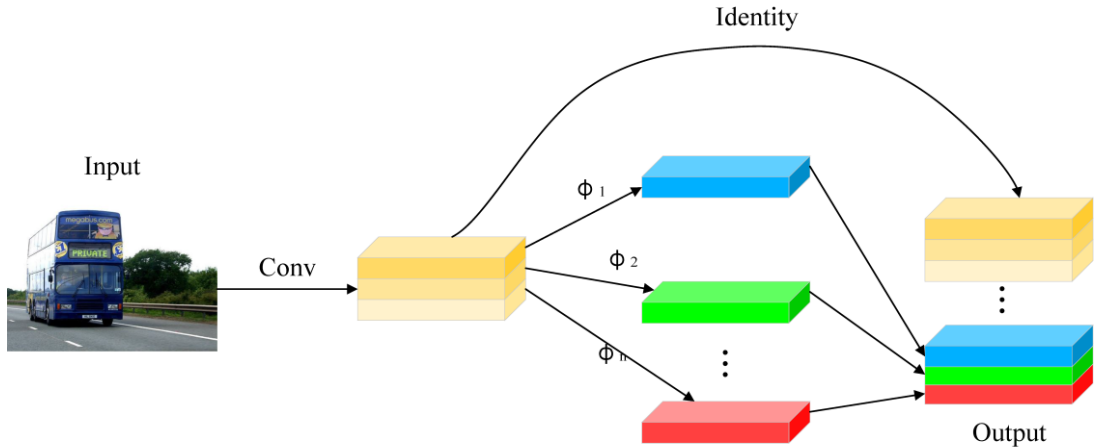

Figure 3 The Ghost convolutional

## 2.2.2. Ghost-BiFPN

In this section, a lightweight neck network named Ghost-BiFPN is designed. YOLOv7 utilizes the combination of FPN and PAN [23] for feature fusion, which cannot guarantee the detection accuracy in different resolution images, leading to a poor performance in detecting of vehicles with varying shapes and sizes. Thus, Ghost-BiFPN is considered as the solution that combines the advantages of YOLOv5 and YOLOv8 neck networks and adds BiFPN [24] and Ghost convolution to reinforce the feature fusion capability of the network and the robustness of the feature map.

Compared to FPN and PAN, BiFPN is specific for efficient bidirectional cross-scale connections and weighted feature fusion. As shown in Figure 4, three feature maps of different sizes are inputed from the backbone network. The features of P3 and P1 are fused directly to the output, which have only one output edge in both stages and has little effect on merging different feature information. An additional edge is added to the output node at the layer of P2, represented by the red line, to fuse more features without increasing the computational cost. Finally, the connection between the input layer of P3 and P2 and the output layer of P1 is added to obtain a higher level of feature fusion.

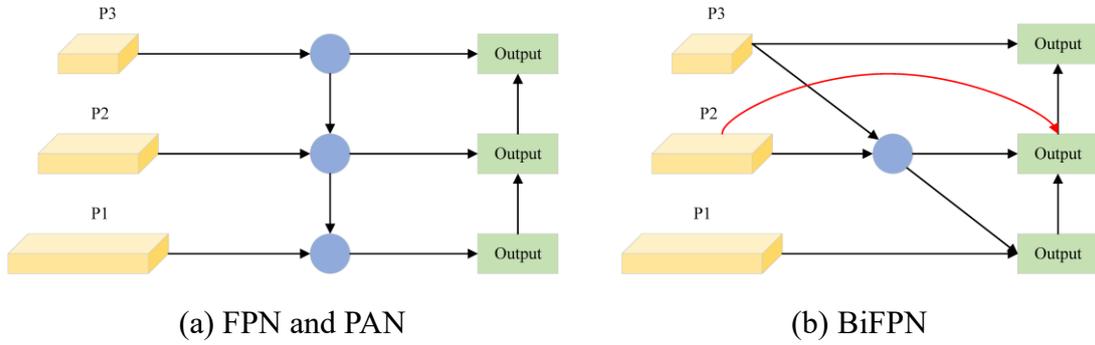

(a) FPN and PAN            (b) BiFPN

Figure 4 The BiFPN structure

When fusing features of different resolutions, it is common practice to adjust them to the same resolution and then sum them up. However, the weights of their output features are usually different due to the variation of input features. Therefore, an extra weight is added to each input of BiFPN to make the network learn the importance of each input feature. Fast normalized fusion is ultilized to adjust the weights in BiFPN, as shown in equation 6:

$$O = \sum_i \frac{w_i}{\varepsilon + \sum_j w_j} \cdot I_i \tag{6}$$

where $w_i$ is a learnable weight and $w_i \geq 0$ is guaranteed by applying a SiLU after each $w_i$, and $\varepsilon = 0.0001$ is to prevent the value from being instability. Similar to the softmax-based fusion approach, each value of the normalized weights is between 0 and 1. The output feature map of each module of BiFPN can be represented as:

$$P_1^{out} = C(\frac{w_1' \cdot P_1^{in} + w_2' \cdot R(P_2^{td})}{w_1' + w_2' + \varepsilon}) \tag{7}$$

$$P_2^{out} = C(\frac{w_1' \cdot P_2^{in} + w_2' \cdot P_2^{td} + w_3' \cdot R(P_1^{out})}{w_1' + w_2' + w_3' + \varepsilon}) \tag{8}$$

$$P_3^{out} = C(\frac{w_1' \cdot P_3^{in} + w_2' \cdot R(P_2^{out})}{w_1' + w_2' + \varepsilon}) \tag{9}$$

where the output feature map at P1, P2 and P3 are denoted by $P_1^{out}$, $P_2^{out}$ and $P_3^{out}$ respectively. The $P_1^{in}$, $P_2^{in}$ and $P_3^{in}$ are denoted the input feature map at P1, P2, and P3 respectively, $C$ is denoted convolution operation, $w_1^{'}$, $w_2^{'}$ and $w_3^{'}$ represent the weights of each layer and $R$ is resizing operation. The intermediate feature $P_2^{td}$ can be represented as:

$$S_2^{td} = C\left(\frac{w_1 \cdot S_2^{in} + w_2 \cdot R(S_3^{in})}{w_1 + w_2 + \varepsilon}\right) \tag{10}$$

where $w_1$ and $w_2$ are the corresponding weights.

### 2.2.3. Ghost Decoupled Head

The detection head of YOLOv7-tiny uses the same coupled head as YOLOv5. However, in object detection, the classification and regression tasks focus on different information, and the coupled head cannot make them focus on the information they each need, so the decoupled head is used to decompose each branch to improve detection accuracy.

In this study, we propose a lightweight Ghost decoupled head by combining the decoupled head of YOLOX [25] and the efficient decoupled head of YOLOv6 [26]. Specifically, the number of 3×3 convolutional layers in YOLOX are reduced to one and replace the standard 3×3 convolutional layers with Ghost convolution, which effectively reduces the computational cost. The structure of the Ghost decoupled head is shown in Figure 5.

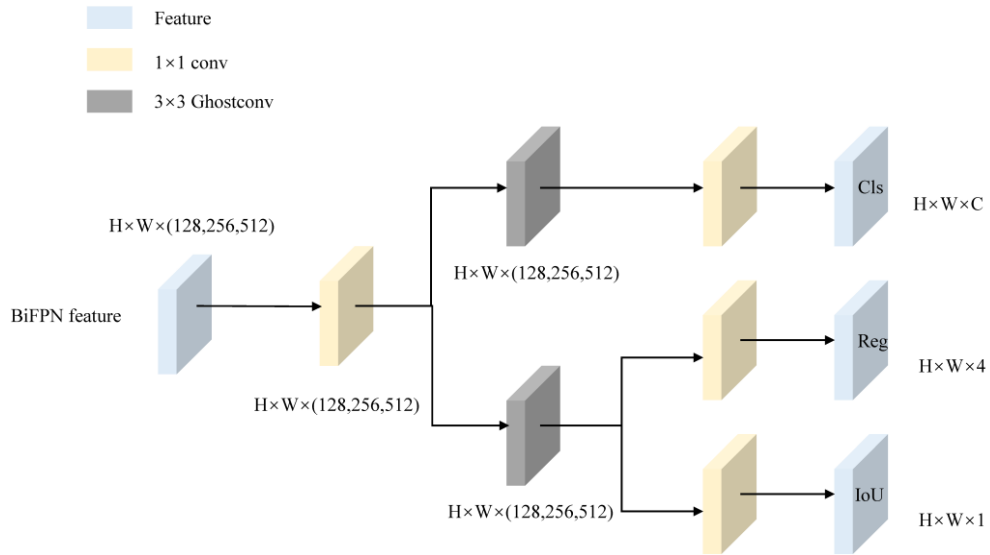

Figure 5 The structure of the Ghost decoupled head

## 2.2.4. Coordinate attention mechanism

In complex scenes, the detection of massive objects and the complexity of locations lead to the loss of information in the network. Hence, a coordinate attention mechanism [27] is added to the network to make the network pay more attention to the detected target and inhibit irrelevant information. SE attention mechanisms evidence that channel attention has a good effect on improving model performance through ignoring location information, which is also considered necessary for spatial feature maps by coordinate attention mechanisms.

The coordinate attention mechanism is considered as a computational unit that aims to enhance feature representation in the network, which encodes channel relations and long-term dependencies through precise location information. A global average pooling is decomposed into two one-dimensional feature encoding operations by two pooling kernels. The first channel is encoded with a pooling kernel of size $(H,1)$ along the horizontal coordinates for the channel. For a given input $X$, the output of the $c_{th}$ channel at height $h$ is calculated in Equation 11:

$$z_c^h(h) = \frac{1}{W} \sum_{0 \leq i < W} x_c(h,i) \quad (11)$$

The other channel is encoded with a pooling kernel of size $(1,W)$ along the vertical coordinates of the channel. The output of the $c_{th}$ channel at width $w$ is calculated in Equation 12:

$$z_c^w(w) = \frac{1}{H} \sum_{0 \leq j < H} x_c(j,w) \quad (12)$$

The above two transformations aggregate features along two spatial directions to obtain the corresponding direction-aware feature maps. The global perceptual field can be well obtained by these two transformations and encodes the precise location information. To utilize the representations generated by the transformations, an extra transformation is proposed, where the above transformations are subjected to a stacking operation and then transformed with a 1 × 1 convolutional transform function $F_1$ as calculated in equation 13:

$$\mathbf{f} = \delta(F_1([z^h, z^w])) \quad (13)$$

where $[z^h, z^w]$ denotes the stacking operation along the vertical and horizontal directions, $\delta$ is the nonlinear activation function, and $\mathbf{f}$ is the intermediate feature map encoded along the horizontal and vertical directions. $\mathbf{f}$ is decomposed into two independent tensors along the spatial dimension. The number of channels is converted to a size consistent with the number of channels of the input $X$ using two 1 × 1 convolutions. The conversion process is shown in Equation 14 and Equation 15:

$$g^h = \sigma(F_h(\mathbf{f}^h)) \quad (14)$$

$$g^w = \sigma(F_w(f^w)) \qquad (15)$$

where $F_h$ and $F_w$ represent the 1 × 1 convolution, $\sigma$ is the sigmoid activation function, and the obtained $g^h$ and $g^w$ are used as the attention weights. Finally, the output of the coordinate attention module is shown in Equation 16, and the structure of the coordinate attention module is shown in Figure 6.

$$y_c(i,j) = x_c(i,j) \times g_c^h(i) \times g_c^w(j) \qquad (16)$$

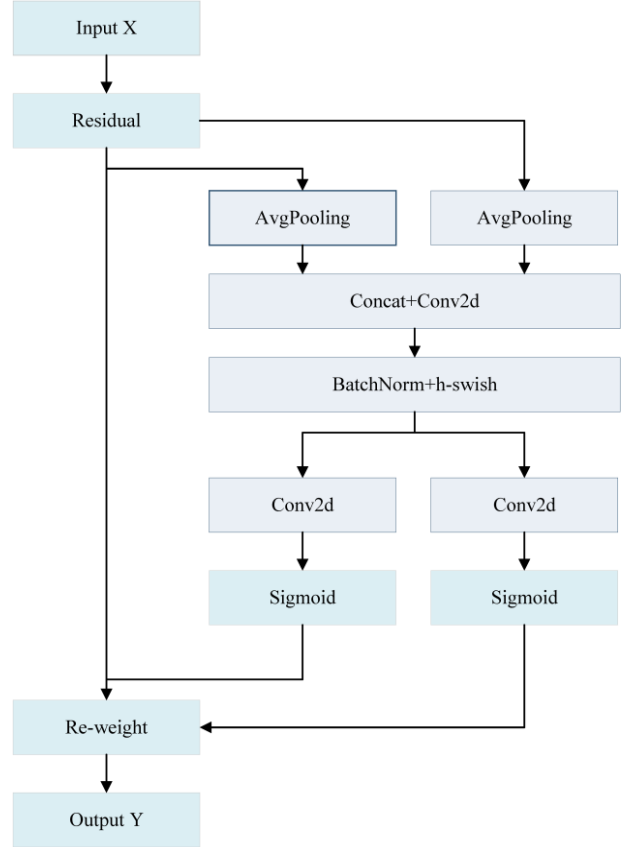

Figure 6 The coordinate attention mechanism structure

## 2.2.5. Loss function improvement

The loss function of YOLOv7-tiny consists of three parts: the confidence loss $l_{obj}$, the classification loss $l_{cls}$, and the bounding box loss $l_{box}$, as shown in equation 17. The bounding box loss consists of bounding box position loss and bounding box prediction loss. The calculation formula of the loss function can be expressed as:

$$Loss = l_{obj} + l_{cls} + l_{box} \qquad (17)$$

The confidence loss can be denoted as:

$$l_{obj} = \sum_{i=0}^{s^2}\sum_{j=0}^{B} I_{ij}^{obj}(\hat{C}_i \log(C_i) + (1-\hat{C}_i)\log(1-\hat{C}_i)) - \\ \lambda_{noobj}\sum_{i=0}^{s^2}\sum_{j=0}^{B} I_{ij}^{noobj}(\hat{C}_i \log(C_i) + (1-\hat{C}_i)\log(1-\hat{C}_i))  \quad (18)$$

The classification loss can be calculated by equation 19:

$$l_{cls} = \sum_{i=0}^{s^2} l_{ij}^{obj} \sum_{c \in classes} (\hat{P}_i(c)\log(P_i(c)) + (1-\hat{P}_i(c))\log(1-P_i(c))) \quad (19)$$

The CIoU [28] loss function is used for regression prediction of the bounding box:

$$Loss_{IoU} = 1 - IoU, IoU = \frac{|A \cap B|}{|A \cup B|} \quad (20)$$

$$Loss_{CIoU} = 1 - IoU + \frac{\rho^2_{(b,b^{gt})}}{d^2} + \alpha v \quad (21)$$

where the parameters $A$ and $B$ represent the areas of the ground-truth and prediction bounding boxes; $b, b^{gt}$ represent the centroids of the prediction and ground-truth bounding boxes, respectively; $\rho$ represents the Euclidean distance between the two centroids; $d$ is the diagonal distance of the smallest enclosing region that contains both the prediction and ground-truth bounding boxes; $\alpha$ is the weight function; $v$ is used to measure the similarity of the aspect ratios. The expression of $\alpha$ and $v$ is as follows:

$$\alpha = \frac{v}{(1 - IoU) + v} \quad (22)$$

$$v = \frac{4}{\pi^2}(arctan\frac{w^{gt}}{h^{gt}} - arctan\frac{w}{h})^2 \quad (23)$$

Although the CIoU loss function solves the problem of gradient disappearance that may be caused by the IoU loss function during training by considering factors such as aspect ratio, there is still room for optimization. The WIoU [29] loss function evaluates the quality of the anchor box by replacing IoU with an outlier degree through a dynamic non-monotonic focusing mechanism and proposes a wise gradient gain allocation strategy that significantly improves the performance. The equation for the WIoU loss function is shown below:

$$Loss_{WIoU} = r \cdot \exp(\frac{\rho^2_{(b,b^{gt})}}{(d^2)^*}) \cdot (1 - IoU) \quad (24)$$

$$r = \frac{\beta}{\delta\alpha^{\beta-\delta}}, \beta = \frac{(1-IoU)^*}{1-IoU} \in [0, +\infty) \quad (25)$$

where $*$ indicates that the dimensions of the smallest bounding box are detached from the computational graph; $\beta$ is the outlier degree; $\alpha$ and $\delta$ are the hyperparameters, where we take $\alpha = 1.9$ and $\delta = 3$.

Compared with CIoU loss function, the WIoU loss function uses an outlier degree to dynamically adjust the gradient gain of anchor box to improve the gradient gain of anchor box as much as possible during the training to achieve the improvement of detection accuracy.

## 3、Experiment

### 3.1. Dataset

The differences between vehicles and the diversity of environment have significant influences on detection speed and detection accuracy. To validate the efficiency of Ghost-YOLOv7, tests are carried on three public datasets, namely Pascal VOC2007+2012 [30], KITTI [31], and BIT-Vehicle [32].

### 3.2. Experimental environment and evaluation indicators

The hardware environment for the experiments is as follows: CPU is Intel Core i5-11400; GPU is Nvidia GeForce RTX 3060Ti, 8G display memories. The software environment for the experiments is as follows: The operating system is Ubuntu 18.04, the CUDA version is 11.3, the programming language is Python 3.8, and the deep learning framework is Pytorch 1.12.1. For the training details, the optimizer is used with Adam; the initial learning rate is 0.001; the number of training epochs is set to 100; the batch size is set to 16; the input size of the image is set to 640×640, and the IoU threshold is set to 0.5. To speed up the convergence of the models, each model uses weights pre-trained on ImageNet.

The mean average precision(mAP), frames per second(FPS), number of parameters, and the floating point operations (FLOPs) are the key indexes for model validation. The mAP is frequently used to evaluate the model's accuracy in object detection. It depends on the average precision(AP) and IoU. While AP depends on precision and recall, they can be expressed as:

$$P(presicion) = \frac{TP}{TP+FP} \tag{26}$$

$$R(recall) = \frac{TP}{TP+FN} \tag{27}$$

$$AP = \int_0^1 P(R)dR \tag{28}$$

$$mAP = \frac{1}{n}\sum_{i-1}^{n}\int_0^1 P(R)dR \tag{29}$$

where $TP$ denotes true positive, $FP$ denotes false positive, $FN$ denotes false negatives, and $n$ is the number of categories. FPS represents the model's speed on the platform, while the number of parameters and the FLOPs measure the computational cost.

## 3.3. Results of ablation experiments

To validate the model, ablation experiments are performed on Pascal VOC series. There are 20 categories in the dataset of the Pascal VOC series, but this paper focuses on only five categories commonly found in traffic scenarios: bicycles, motorcycles, buses, cars, and people. Since the number of one single dataset is not large enough, this paper combines VOC2007 and VOC2012 to obtain a total of 16,441 images, which are divided into the training set, validation set, and test set according to 8:1:1.

Table 1 shows the compositions of models generated by different improvement methods, and Table 2 shows the results of the ablation experiments. In the case of model 1, the number of parameters and computation of the model and the model size is reduced by nearly four times compared with the original model. In the case of model 2, the standard convolution of the backbone network is replaced by Ghost convolution to lighten the network further. In the case of model 3, it can be seen that Ghost-BiFPN increases the number of parameters but significantly improves the detection accuracy. In the case of model 4, the GDH sacrifices some of the detection speed to improve the detection accuracy. In the case of model 5 and model 6, coordinate attention mechanism and WIoU loss function further improves detection accuracy without increasing parameters. Compared with the original model, parameters are reduced by 37.3%, 29.8% less computation, 35.1% less model size, and 1.1% more mAP, and the improved model has a detection speed of 428 FPS on the test dataset when the batch size(bs) is set to 16.

Table 1 The combination of different improvement methods results.

| **Model** | 0.5width | Ghost-Backbone | Ghost-BiFPN | GDH | CA | WIoU |
|---|---|---|---|---|---|---|
| Model 1 | √ | | | | | |
| Model 2 | √ | √ | | | | |
| Model 3 | √ | √ | √ | | | |
| Model 4 | √ | √ | √ | √ | | |
| Model 5 | √ | √ | √ | √ | √ | |
| Model 6 | √ | √ | √ | √ | √ | √ |

Table 2 Comparison of ablation experiments of the overall model.

| **Model** | **Params(M)** | **FLOPs(B)** | **mAP@0.5(%)** | **Weight(MB)** | **FPS(bs=16)** |
|---|---|---|---|---|---|
| YOLOv7-tiny | 6.01 | 13.1 | 80.5 | 11.7 | 401 |
| Model 1 | 1.51 | 3.4 | 74.6 | 3.1 | 500 |
| Model 2 | 1.33 | 3.0 | 74.2 | 2.8 | 525 |
| Model 3 | 2.00 | 4.4 | 78.1 | 4.0 | 481 |
| Model 4 | 3.76 | 9.2 | 79.6 | 7.6 | 448 |
| Model 5 | 3.77 | 9.2 | 80.8 | 7.6 | 428 |
| Model 6 | 3.77 | 9.2 | 81.6 | 7.6 | 428 |

## 3.4. Comparison analysis of algorithms

### 3.4.1. performance on the Pascal VOC2007+2012 dataset

The VOC 2007 and 2012 datasets are recognized as challenging object detection datasets. There are 9963 images in the VOC2007 dataset, and 11530 images in the VOC2012 dataset. This section compares the performances of different algorithms on this VOC2007+2012 dataset. As shown in Table 3, Ghost-YOLOv7 achieves the highest detection speed of 428 FPS. Although the number of parameters and computation of the model is more prominent than YOLOv5n, Ghost-YOLOv7 also has the highest detection accuracy among the lightweight models. The detection results are shown in Figure 7. Comparing YOLOv7-tiny, it can be seen that Ghost-YOLOv7 is superior and more robust for multi-target detection.

Table 3 Comparison results of different algorithms on the VOC2007+2012 dataset.

| Model | Params(M) | FLOPs(B) | mAP@0.5(%) | Weight(MB) | FPS(bs=16) |
|---|---|---|---|---|---|
| YOLOv7-tiny | 6.01 | 13.1 | 80.5 | 11.7 | 401 |
| YOLOv5n | 1.76 | 4.1 | 77.3 | 3.8 | 385 |
| YOLOv3-tiny | 8.67 | 12.9 | 63.1 | 16.6 | 345 |
| Faster R-CNN | 137.07 | 626.2 | 78.2 | 521.0 | 14 |
| Ghost-YOLOv7 | 3.77 | 9.2 | 81.6 | 7.6 | 428 |

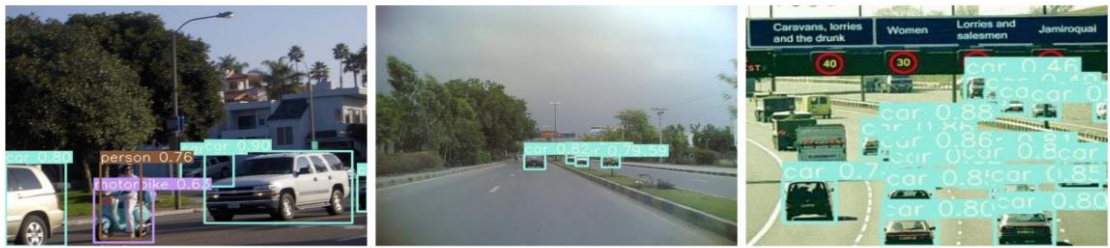

(a) Ghost-YOLOv7

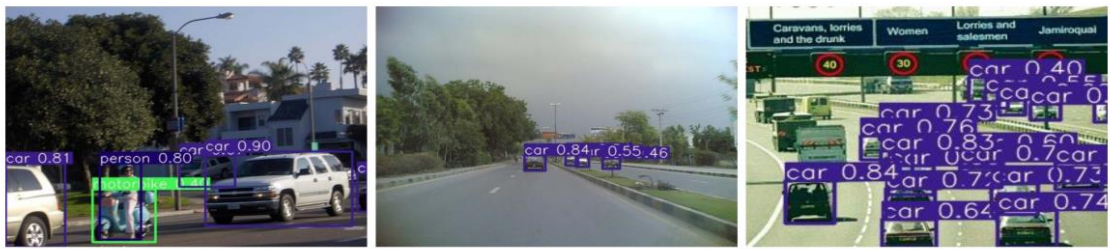

(b) YOLOv7-tiny

Figure 7 Comparison of the detection results of Ghost-YOLOv7 and YOLOv7-tiny on the VOC2007+2012 dataset.

## 3.4.2. performance on the KITTI dataset

The KITTI dataset installs vehicle cameras to capture image data from urban, rural, and highway scenarios. The dataset consists of 7481 images and contains five categories: car, people, cyclist, van, and truck, but in the evaluation phase, it officially only considers three categories of car, people, and cyclist. Since the ground truth of the test set is not officially provided, this paper divides the training and test sets into 3712 and 3769 images. The evaluation of the KITTI dataset is classified as easy, moderate, and complicated according to the bounding box's height and the occlusion level. As shown in Table 4, Ghost-YOLOv7 achieved 88.4% mAP on the KITTI dataset, which is a 1% improvement compared to YOLOv7-tiny, and it also performs better at all three different difficulty levels. Ghost-YOLOv7 is more accurate in detecting vehicles facing small targets and different sizes.

Table 4 Comparative analysis of different algorithms on the KITTI dataset. E stands for Easy, M stands for Moderate, and H stands for Hard.

| Model | AP(%) | | | | | | | | | mAP(%) | FPS(bs=16) |
|---|---|---|---|---|---|---|---|---|---|---|---|
| | Car | | | Person | | | Cyclist | | | | |
| | E | M | H | E | M | H | E | M | H | | |
| YOLOv7-tiny | 97.6 | 90.0 | 88.5 | 87.3 | 78.5 | 77.1 | 89.5 | 87.5 | 80.1 | 87.4 | 401 |
| YOLOv5n | 97.1 | 89.6 | 80.5 | 84.3 | 76.2 | 74.6 | 87.9 | 79.5 | 78.8 | 85.7 | 385 |
| YOLOv3-tiny | 90.1 | 85.4 | 76.4 | 81.5 | 73.3 | 66.4 | 83.1 | 75.2 | 74.4 | 79.6 | 345 |
| Faster R-CNN | 97.1 | 89.8 | 79.6 | 85.1 | 76.7 | 75.2 | 88.2 | 79.6 | 78.7 | 86.1 | 14 |
| Ghost-YOLOv7 | 97.4 | 90.1 | 88.7 | 87.3 | 78.7 | 77.4 | 89.7 | 87.7 | 80.2 | 88.4 | 428 |

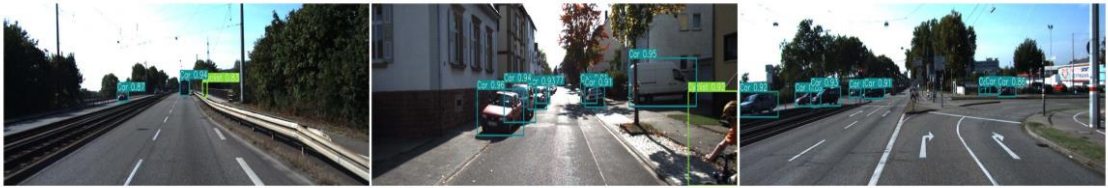

(a) Ghost-YOLOv7

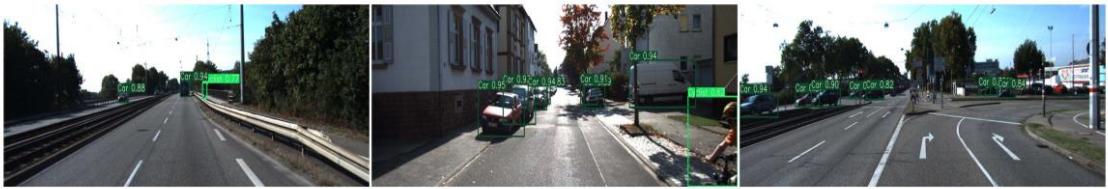

(b) YOLOv7-tiny

Figure 8 Comparison of detection results of Ghost-YOLOv7 and YOLOv7-tiny on the KITTI dataset.

### 3.4.3. performance on the BIT-Vehicle dataset

The BIT-Vehicle dataset is produced by the Beijing Institute of Technology. The dataset contains 9850 images, most of which have only one to two vehicles, and the dataset as a whole is relatively simple. The dataset is divided into six categories: bus, microbus, minivan, sedan, SUV, and truck. In this paper, the dataset is divided into the training set, validation set, and test set according to 8:1:1, and the experimental results on BIT-Vehicle dataset are shown in Table 5. The proposed model in this paper achieves 99.1% mAP, which is a good improvement in all aspects compared to the original model. As shown in Figure 9, Ghost-YOLOv7 has a higher detection effect and better detection in the face of obscured targets.

Table 5 Experimental results of different algorithms on the BIT-Vehicle dataset.

| Model | Params(M) | FLOPs(B) | mAP@0.5(%) | Weight(MB) | FPS(bs=16) |
|---|---|---|---|---|---|
| YOLOv7-tiny | 6.01 | 13.1 | 98.5 | 11.7 | 401 |
| YOLOv5n | 1.76 | 4.1 | 98.0 | 3.8 | 385 |
| YOLOv3-tiny | 8.67 | 12.9 | 97.8 | 16.6 | 345 |
| Faster R-CNN | 137.07 | 626.2 | 98.1 | 521.0 | 14 |
| Ghost-YOLOv7 | 3.77 | 9.2 | 99.1 | 7.6 | 428 |

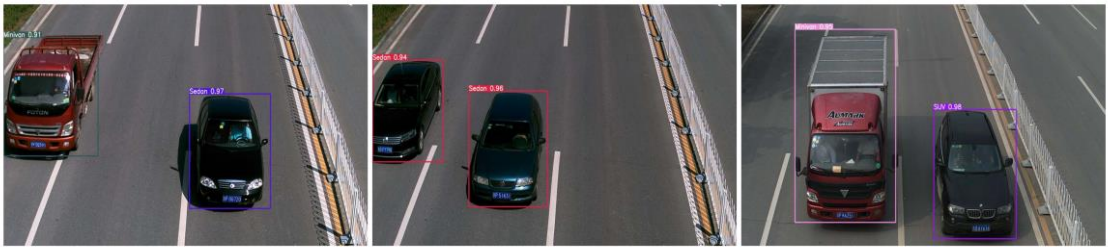

(a) Ghost-YOLOv7

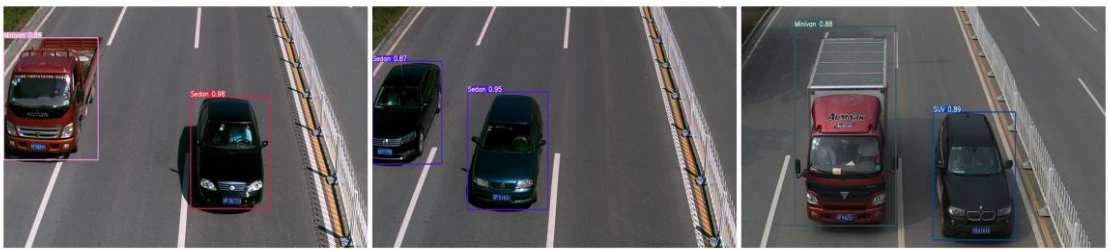

(b) YOLOv7-tiny

Figure 9 Comparison of the detection results of Ghost-YOLOv7 and YOLOv7-tiny on the BIT-Vehicle dataset.

# 4、Conclusion

Based on YOLOv7-tiny，a lightweight vehicle detection model named Ghose-YOLOv7 is proposed in this study. The model scales the width multiple to 0.5 and replace the standard convolution of the backbone network by Ghost convolution, a Ghost-BiFPN neck network is designed to enhance the feature fusion capability; then a lightweight Ghost decoupling head is added to improve the convergence speed and detection effect; finally, a coordinate attention mechanism and WIoU loss function are introduced to enhance the detection accuracy further. The experimental results on different datasets, namely VOC2007+2012, KITTI, and BIT-Vehicle datasets show that Ghost-YOLOv7 dramatically reduces the number of parameters of the model while improves the detection speed. Moreover, it has higher detection accuracy and lower computational cost than other detection algorithms in most cases. It can be accredited for effective implementation of the algorithm in practical application. Further work can be considered to enhance the model's detection accuracy and deploy the improved model to resource-constrained on-board edge platforms.

## CRediT authorship contribution statement

**Bo Li:** Resources, Writing – review & editing, Supervision. **YiHua Chen:** Investigation, Validation, Data curation. **Hao Xu:** Conceptualization, Visualization, Writing – original draft, Software.

## Declaration of competing interest

The authors declare that they have no known competing financial interests or personal relationships that could have appeared to influence the work reported in this paper.

## Data availability

Data will be made available on request.

## Acknowledgments

This research was sponsored by the National Natural Science Foundation of China (Grant No.: 52005168), the Initial Scientific Research Foundation of Hubei University of Technology (BSQD2020005), and the Open Foundation of Hubei Key Lab of Modern Manufacture Quality Engineering (KFJJ-2020009).